\documentclass[letterpaper]{article}
\usepackage{iccc}
\usepackage[english]{babel}
\usepackage{todonotes}
\usepackage{glossaries}
\usepackage{times}
\usepackage{helvet}
\usepackage{courier}
\usepackage[sort]{natbib}
\usepackage{booktabs, array}
\usepackage{hyperref}
\usepackage[OT1]{eulervm}

\newcommand{\citepos}[1]{
\citeauthor{#1}'s (\citeyear{#1})}

\newcommand{\citeposs}[1]{
\citeauthor{#1}' (\citeyear{#1})}

\usepackage{mathrsfs}
\usepackage{amsmath}
\usepackage{amsthm}
\usepackage{amssymb}
\usepackage[nameinlink]{cleveref}
\usepackage{thmtools}
\usepackage{setspace}
\usepackage{enumitem}

\DeclareMathAlphabet\mathbfcal{OMS}{cmsy}{b}{n}

\declaretheoremstyle[bodyfont=\slshape, notefont=\textbf, notebraces={}{} ,headformat=\NAME~\NUMBER~-\NOTE]{ilshape}

\declaretheoremstyle[bodyfont=\slshape, notebraces={}{}]{slshape}

\declaretheorem[style=slshape,name=Theorem]{theorem}

\declaretheorem[style=ilshape,name=Definition]{definition}

\newenvironment{proofs}[1][]
{
    \providecommand{\theproof}
        {Proof\cref*{#1}}
    \begin{proof}[\rm\bf Proof 
    ]}
{\end{proof}}

\newlist{deflist}{enumerate}{1}
\setlist[deflist]{label=(\alph*),
                  ref=\thedefinition.(\alph*)}

\newlist{thmlist}{enumerate}{1}
\setlist[thmlist]{label=(\alph*),
                  ref=\thetheorem.(\alph*)}

\newlist{thmlistinline}{enumerate*}{1}
\setlist[thmlistinline]{label=(\alph*),
                  ref=\thetheorem.(\alph*)}

\newlist{prooflistinline}{enumerate*}{1}
\setlist[prooflistinline]{label=(\alph*),
                  ref=\theproof.(\alph*)}

\DeclareMathOperator*{\content}{content}
\DeclareMathOperator*{\range}{range}

\DeclareMathOperator{\converges}{{\downarrow}}
\DeclareMathOperator{\diverges}{{\uparrow}}

\crefname{definition}{Def}{Defs}
\crefname{theorem}{Thm}{Thms}
\crefname{lemma}{Lem}{Lems}
\crefname{proposition}{Prop}{Props}
\crefname{deflisti}{Def}{Defs}
\crefname{thmlisti}{Thm}{Thms}
\crefname{thmlistinlinei}{Thm}{Thms}
\crefname{prooflistinlinei}{}{}

\title{Towards a Formal Creativity Theory: \\ Preliminary results in Novelty and Transformativeness}

\author{
Luís Espírito Santo\\
CMS -- CISUC, \\
University of Coimbra, Portugal \\
lesanto@dei.uc.pt\\
CCLab -- AILab, \\ 
Vrije Universiteit Brussel, Belgium \\
luis.espirito.santo@vub.be\\ 
\And
Geraint Wiggins\\
CCLab -- AILab \\
Vrije Universiteit Brussel, Belgium \\
Cognitive Science Group\\
Queen Mary University of London, UK\\
geraint.wiggins@vub.be\\
\And
Amílcar Cardoso\\
Centre for Informatics and Systems (CISUC), \\
University of Coimbra, Portugal \\
amilcar@dei.uc.pt\\
}

\newacronym{acc}{ACC}{International Association for Computational Creativity}
\newacronym{ae}{AE}{Auto-Encoder}
\newacronym{bce}{BCE}{Binary Cross-Entropy}
\newacronym{ai}{AI}{Artificial Intelligence}
\newacronym{cc}{CC}{Computational Creativity}
\newacronym{ct}{CT}{Convergent Thinking}
\newacronym{cce}{CCE}{Categorical Cross-Entropy}
\newacronym{cct}{CCT}{Computational Creativity Theory}
\newacronym{clt}{CLT}{Computational Learning Theory}
\newacronym{csf}{CSF}{Creative Systems Framework}
\newacronym{dl}{DL}{Deep Learning}
\newacronym{dt}{DT}{Divergent Thinking}
\newacronym{flt}{FLT}{Formal Learning Theory}
\newacronym{fct}{FCT}{Formal Creativity Theory}
\newacronym{gan}{GAN}{Generative Adversarial Networks}
\newacronym{gru}{GRU}{Gated Recurrent Units}
\newacronym{legs}{LEGS}{Learning-Endowed Generative Systems}
\newacronym{lstm}{LSTM}{Long-Short Term Memory}
\newacronym{mae}{MAE}{Mean Absolute Error}
\newacronym{ml}{ML}{Machine Learning}
\newacronym{mle}{MLE}{Maximum Likelihood Estimator}
\newacronym{mse}{MSE}{Mean Squared Error}
\newacronym{rbm}{RBM}{Restricted Boltzman Machines}
\newacronym{rnn}{RNN}{Recurrent Neural Networks}
\newacronym{silitf}{SILIT}{Scientist Identifying Languages in Texts}
\newacronym{vae}{VAE}{Variational Auto-Encoders}
\newacronym{vq-vae}{VQ-VAE}{Vector Quantised-Variational Auto-Encoder}
\newacronym{vq-gan}{VQ-GAN}{Vector Quantised-Generative Adversarial Networks}
\newacronym{re}{r.e.}{recursively enumerable}

\usepackage{titlesec}

\renewcommand*{\thesubsubsection}{\arabic{subsubsection}.}
\counterwithin*{subsubsection}{subsection}
\renewcommand\thesubsubsection{\thesubsection.\arabic{subsubsection}}%
\titleformat{\subsubsection}{\normalfont\bfseries}{\thesubsubsection}{1em}{}
  
\begin{document} 
\maketitle
\begin{abstract}
\begin{quote}
Formalizing creativity-related concepts has been a long-term goal of Computational Creativity. To the same end, we explore Formal Learning Theory in the context of creativity. We provide an introduction to the main concepts of this framework and a re-interpretation of terms commonly found in creativity discussions, proposing formal definitions for novelty and transformational creativity. This formalisation marks the beginning of a research branch we call Formal Creativity Theory, exploring how learning can be included as preparation for exploratory behaviour and how learning is a key part of transformational creative behaviour. By employing these definitions, we argue that, while novelty is neither necessary nor sufficient for transformational creativity in general, when using an inspiring set, rather than a sequence of experiences, an agent actually requires novelty for transformational creativity to occur.
\end{quote}
\end{abstract}

\section{Introduction}

Since its origins in the 1960s, the theory known as the \gls{flt} was motivated by the task of providing natural language capabilities to machines. It has equally been used to explore long-lasting debates on children's intellectual development 
or even to structure philosophical fields such as epistemology. It is a nonstatistical research stem within \gls{clt} which facilitates the examination of limitations pertaining to learning under theoretical conditions, such as infinite data, time, and space. As such, both language acquisition and scientific discovery are traditionally employed and vastly explored for the purpose of illustrating and inspiring \gls{flt}.

Some authors, such as \cite{Koestler2014-hp}, argue that creativity has a major role in tasks like language acquisition and scientific discovery, often seen solely as learning. However, others find this hard to accept, as \citet[p.~69]{Ritchie2007-ii} suggests: ``Mathematics,
science, or engineering are rarely classed as creative unless they are done exceptionally well. This bias does not seem helpful in a rigorous attempt to pin down the notion of creativity, particularly when applied to machines. Although there is still a tendency within AI to tacitly accept this intellectual apartheid of creative versus non-creative activities, it would be better if we could be more neutral in our formal characterisation of creative actions.''

We believe that some ideas in \gls{flt} could have a useful purpose for creativity studies, especially in structuring and formalizing matters. Our exploration of this theory is an attempt at formalizing concepts in \gls{cc}, following a series of similar attempts that are based on other theoretical fields such as: the early developments by \citet{Wallas1926-og}, \citet{Rhodes1961-go} and \citet{Guilford1950-fl}; \citepos{Boden1994-et} ideas and their subsequent examination by others such as \cite{Bundy1994-ef} and \cite{Ritchie2006-ye}; \citeposs{Wiggins2019-kv} progress at bringing extra rigour to \citepos{Boden2004-ha} proposal; \citepos{Schmidhuber2010-zy} project to formalise creativity; \citepos{Colton2011-rt} proposal of Computational Creativity Theory;  \citepos{Ventura2011-dz} adaptation of the ``No Free Lunch Theorem''; Shannon's Information Theory use by \cite{Varshney2019-ta,Varshney2020-li} and the approach to Kolmogorov Complexity by \cite{Mondol2021-qa}.

In this paper, our goal is three-fold: 1) to introduce the ideas behind \gls{flt}; 2) to provide formal definitions for some creativity-related terminology, inspired in \gls{flt}; 3) to show how these formal definitions allow to explore and demonstrate results about creativity, by exploring the effective role of novelty in transformational creativity \citep{Boden2004-ha}.

\section{The Endless Explanation of ``Grey Music''}\label{sec:motivation}

In this section, we present an anecdote, a scenario, that sets the stage for our explanation of \gls{flt}, by illustrating the foundational concepts within the context of a creative domain and highlighting some important aspects. 
Since \gls{flt} is agnostic and applicable across domains, this anecdote could pertain to any creative domain and still illustrate the theory as effectively. Besides, adapting it for other domains should be easily achievable. We chose the music domain and titled our anecdote: The Endless Explanation of ``Grey Music''.

Our story begins with a vinyl shop assistant and his client\footnote{\gls{flt} is agnostic in terms of what is considered a data provider (client) and a learning agent (assistant). One can think of any data-provider in place of the client such as a critic, an audience, nature, or the assistant himself, and any data-consuming/hypothesis-producing entity in place of the assistant like a scientists, a machine, a team of people. While important for generalization, this flexibility can be overly complex for this introduction.}, as they embark on a journey around a musical genre known as \textit{grey music}, in a world where intensional definitions are avoided. 
The anecdote revolves around the client's  extensional description of what she wants for a song and the assistant's efforts to understand this musical genre, which is as mysterious to the assistant as it is clear to the client. 

After acknowledging that the assistant has no idea of what \textit{grey music}, the musical genre, is, the client proceeds to indicate songs: `You know, ``Le Freak'' by Chic. That is \textit{grey music}!'\footnote{Given this single example the reader might already feel the urge to trying to unravel what \textit{grey music} means. This illustrates human's predisposition to create these structures.}. 
The client is determined to share every single \textit{grey music} song, even if there are infinitely many, to ensure the assistant gets it exactly correctly, no more no less, the same way she does. Therefore, she will exhaustively exemplify \textit{grey music}, and every time she mentions a song, the assistant may refine his understanding of the term, adjusting his conceptualization of \textit{grey music} and creating different sets of rules to represent their evolving comprehension.

The never-ending part of this story arrives when we understand that if this genre is big enough, i.e., infinite, and the motivations of the client and assistant never fade then they would be able to stay this way forever. 
The client can never be certain whether the assistant have the exactly correct conceptualization of \textit{grey music} simply by analysing any finite amount of songs suggested by the assistant given his current understanding of the term
-- we will term this situation in which the client lacks direct access to the assistant's current conceptualization as the ``evaluator conundrum''. Thus, the client keeps forever providing examples, wholly independent of the assistants's understanding \footnote{Most \gls{flt} results assume the premise of passive learning, where the learner  observes data without the capacity of intervening. Opposedly, active learning allows the learner to experiment and the data will change according to those actions. This dicothomy is different than the one between implicit vs deliberate learning.}. 
Meanwhile, the assistant also is perpetually uncertain about the exaclty correct \textit{grey music}'s essence since he only has access to the examples and nothing more -- to this other condition we will call the ``learner conundrum'' -- and has no capacity to intervene in the data that is provided.
By indefinitely proceeding the same way, they get stuck in an endless loop:
\begin{itemize}
    \item the client keeps on exhaustively exemplifying \textit{grey music} uncertain if the assistant captured it exactly correctly;
    \item the assistant keeps on trying to acquire the exactly correct conceptualiation for the genre of \textit{grey music}, passing through several hypothetical genres, but without ever knowing if has matched the client's idea exactly correctly.
\end{itemize}

Contrasting with practical research, this kind of thought experiments provide us -- as external observers -- direct access to both the client's and shop assistant's minds. Therefore, despite their conundra, we can determine whether they will ever successfully converge to precisely the same genre or not. \gls{flt} takes advantage of this fact to theoretically explore which learning strategies enable assistants to to achieve success within various environments, seeking to understand the nuanced dynamics between different approaches and the contexts in which they are most effective. Additionally, this method also allows for the study of whether there are genres that are impossible to discern apart.  

\section{An Introduction to Formal Learning Theory}\label{sec:flt}

\citet{E_Mark_Gold1967-ps}  presented the first theoretical results that became the foundations of a formal framework to study inductive learning. Building on these results,  \cite{Blum1975-xq}, \cite{Angluin1980-ye}, \cite{Case1983-bj} and \cite{Case2012-vn}, as well as several other authors, developed a set of other definitions and results that describe limits on what one learner can learn. 
We call the collection of such definitions and results, \gls{flt}, which we now introduce. In this introduction, we reproduce a slight adaptation of some of the content in the comprehensive review of \gls{flt} made available by \cite{Jain1999-gh}. For the sake of clarity and ease of reference, we decided to compile this set of definitions under a common name: \gls{silitf}. This represents what we consider to be the most basic framework presented in this area that will serve as the basis for other possibly more intricate adaptations and extensions.

\subsection{Learning Paradigm}

The anecdote presented illustrates five components that can be found in any empirical inquiry paradigm \citep{Jain1999-gh} and are thus formalized in \gls{silitf}:
\begin{itemize}
    \item a class of possible \textbf{realities}:  In our anecdote, this pertains to all potential music genres, abstracted as sets of songs\label{def:paradigm_realities}.
    \item intelligible \textbf{hypotheses}: These refer the way to represent the music genre that the assistant is thinking of at each moment. This corresponds to the set of rules the assistant produces to represent his hypothetical reality\label{def:paradigm_hypotheses}.
    \item the extensive \textbf{data} available about the correct reality: The data in our anecdote are the songs provided by the client. \label{def:paradigm_data}
    \item a \textbf{learner}: This is anything that can turn data into hypotheses, which corresponds to our assistant; and \label{def:paradigm_learner}
    \item \textbf{criteria} for success: In our case, we aim at what we designated as matching ``exactly correctly'', i.e., the assistant's conceptualization perfectly aligns with the client's, including no additional songs nor omissions. Various alternative criteria are explored in different extensions. \label{def:paradigm_criteria}
\end{itemize}

\subsection{Notation and Terminology}

Here, we introduce some notation, primarily drawn from the theory of computation, thoroughly explored by \cite{Rogers1967-xi}. We denote the empty set as $\emptyset$, while $\mathbb{N}$ represents the set of natural numbers. In this paper, $\mathbb{A}$ encompasses the set of all conceivable sentences within a chosen alphabet, which serves as potential inputs for programs $p$ written for a selected Turing-equivalent algorithmic system. We will use $\mathbb{P}$ to denote the set of all those possible programs. Within our notation, $W_p$ signifies the domain of the computable function expressed by program $p$. So, the statement $L = W_p$ reads \textit{$p$ is a (discriminative)\footnote{Discriminative grammars oppose to generative grammars. Discriminative use the domain in their definition and can be used directly to semi-decide what sentences are well-formed. Generative can be used to always output a well-formed sentence.} grammar for set $L$}. Thus, every program $p \in \mathbb{P}$ is a grammar for exactly one set $L\subseteq \mathbb{A}$. 

\citepos{Turing1937-mo} results imply that not all sets of sentences have grammars, but those that do have a grammar always have a Turing machine that lists all its members. 
Then, we say that those sets which have grammars are therefore \gls{re} sets\footnote{The term \gls{re} set is usually applied to sets of natural numbers and not sentences, yet this is equivalent and simpler.}
and we denote the class of such sets by $\mathcal{E}$, and we can thus write $\mathcal{E} = \{W_p \mid p \in \mathbb{P}\}$.


\subsection{Scientists Identifying Languages in Texts}

We designate \gls{silitf} to address what we consider to be the most basic set of definitions among the several introduced by \cite{Jain1999-gh}. It is arguably the most basic framework in \gls{flt}, that can be expanded, adapted, and appropriated to include other kinds of interesting aspects. This framework as well as its variations are different ways to formalise the five components that we find in any empirical inquiry paradigm. Similarly to many other ideas in the theory of computation, the terminology revolves around languages and sentences, to directly model language acquisition.

We will start by formalizing the class of possible realities, whose formalisation was directly borrowed from computability theory. 
Since the learner must be able to point at specific sets of things, for example, songs or sentences, one way is to allow our learner to produce intrinsical definitions for those sets. Using programs we can actually have a finite representation for potentially infinite sets.  
Thus, in this framework, a possible reality is a \gls{re} set\footnote{There are alternative frameworks to learn functions rather than sets. Yet, the latter is more general and interesting for creativity.}, also designated as a language, $L \in \mathcal{E}$. So, in context of \gls{silitf}:

\begin{definition}[Language (Reality)]
     A language or a (possible) reality is a recursively enumerable set, $L \in \mathcal{E}$. A class of languages will be represented by symbol $\mathcal{L} \subseteq \mathcal{E}$.  \label{def:reality}
\end{definition}

Following this strategic choice, the hypotheses, i.e., the way the learner intrinsically represents these realities, can be abstracted using grammars, or in other words, computer programs, $p \in \mathbb{P}$ \footnote{In \gls{flt} these are represented using natural numbers by employing Gödel numbering as an index.}.
These will be the output of the learner when presented with data. Then every possible reality has at least one corresponding hypothesis and every hypothesis as exactly one reality to which it corresponds.

\begin{definition}[Grammar (Hypothesis)]
    \hfill

    \begin{deflist}
        \item \textbf{Grammar}. In this framework, a grammar, a hypothesis or a conjecture is a program, $p \in \mathbb{P}$. \label{def:hypothesis}
        
        \item  \textbf{Language corresponds to a grammar}. A language $L$ is said to correspond to a grammar $p$ if and only if $L = W_p$.\label{def:reality_of_hypothesis}
    \end{deflist}    
\end{definition}

We address now how this framework models the data collected from one specific fixed reality. To faithfully emulate this way of learning, we must ensure that the learner only accesses the correct reality indirectly through extensional data, i.e., we need to model the learner conundrum by creating a singular interface between the learner and the correct reality.

One simple way to represent data is to consider an infinite sequence of positive examples encompassing all members of the set -- an exhaustive representative dataset. In this area, this list-like representation of all the elements of a set is termed as `text', since this would correspond to an actual text with all the sentences that could be written in a language. In the same way, the words in a text do not change based on what a reader is thinking, this structure is also independent of what the learner is outputting and may include repetitions or even blank spaces, points of null data. In our case, it will be an infinite sequence of sentences that belong to the chosen reality, without missing one. We also include an extra symbol, $\#$, that represents the inexistence of data.

\begin{definition}[Text (Data)]\label{def:text}

    \hfill

    \begin{deflist}
    
        \item \textbf{Text}. A text $T$  is a mapping $\mathbb{N} \rightarrow \mathbb{A} \cup \{ \#\}$. 
        This can and will also be equivalently interpreted and noted as an infinite sequence $x_0, x_1, x_2, \ldots$ of sentences where some might be null, $\#$, instead. \label{def:text_text}
        
        \item \textbf{Content of a text}. The sentences appearing in this 
        sequence are denoted $\content(T) = \range(T)-\{\#\}$. \label{def:text_content}
        
        \item \textbf{Text exhaustively exemplifies a language}. A text $T$ is said to exhaustively exemplify a language $L \in \mathcal{E}$ in case $\content(T) = L$. In some literature, this reads $T$ is for $L$.
        
        \item \textbf{Text operators}. Given a text $T$ and $n \in \mathbb{N}$, then $T(n)$ 
        represents the n+1\textsuperscript{th} member of $T$, while $T[n]$ 
        represent the initial finite sequence of length $n$, known as prefix.  \label{def:text_sets_operators}
        
        \item \textbf{Prefix operators}. Prefixes are represented using the following symbols: $\sigma, \tau$. The length of prefix $\sigma$ is 
        denoted $|\sigma|$, the concatenation of prefixes writes $\sigma \diamond \tau$ 
        and the remaining notation and operators for texts (content, prefixes and indexing) 
        defined in \cref{def:text_sets_operators}, also apply to prefixes.\label{def:text_sets_prefix_operators}
        
        \item \textbf{The set of all set prefixes}. The set of all prefixes is written $SEQ = \{T[n]  \mid   T \text{ a text and } n \in \mathbb{N}\}$.
        
    \end{deflist}
\end{definition}

Notice that for each language $L$ there are 
infinite many texts, 
except for the empty set $\emptyset$ which only accepts one single text, 
namely, $T_\emptyset$ such that $\forall_{x\in \mathbb{N}} T_\emptyset(x) = \#$. For example, the following three texts all exhaustively exemplify the reality of the sentences representing even numbers:

$$2,4,6,8,10, 12, 14, 16, 18\ldots$$

$$\#, 2, \#, 4, 4, \#, 6, 6, 6, \# \ldots$$

$$\#, 2, 6, 4, 10, 8, 14, 12, 18\ldots$$

We are now in a condition to formalise the idea of learner, or scientist, as they are called in this area. We want something that given partial finite data outputs a hypothesis.

\begin{definition}[Scientist (Learner)]\label{def:scientists} 

    \hfill
    
    \begin{deflist}
    
        \item \textbf{Scientist}. A scientist is a function $\mathcal{M}: SEQ \rightarrow \mathbb{P}$. \label{def:scientist}
        
        \item \textbf{Scientist conjectures reality}. Let $T$ be a text and $n \in \mathbb{N}$, 
        then a scientist $\mathcal{M}$ is said to conjecture the language $L$ 
        on prefix $T[n]$ if and only if $W_{\mathcal{M}(T[n])} = L$, i.e., if scientist $\mathcal{M}$ outputs a grammar for $L$ when given as input the initial sequence of length $n$ from $T$. 
    \end{deflist}
\end{definition}

It is essential to note that this definition of scientist is notably broad, encompassing even scientists that can never be implemented in computers, i.e., uncomputable scientists. Such wide-ranging definition was crucial to drawing conclusions regarding the computability of learning since it facilitated the comparison of the learning limitations of this broad class of scientists with those limitations encountered when considering exclusively computable scientists. We proceed to present one criterion that assesses the success of a scientist in a given text: identifiability.

\begin{definition}[Identification (Criterion)]\label{def:identification}
Let $\mathcal{M}$ be a scientist and $T$ a text, respectively.

    \begin{deflist}
    
        \item \textbf{Convergence on a text}. $\mathcal{M}(T)\converges = p$, 
        reading $\mathcal{M}$ converges to $p$ on $T$, if and only if 
        for all but finitely many $n \in \mathbb{N}$, we have $\mathcal{M}(T[n]) = p$. 
        If $\mathcal{M}$ doesn't converge to any $p$ then we write $\mathcal{M}(T)\diverges$, reading $\mathcal{M}$ diverges in $T$.\label{def:convergence_text}
        
        \item \textbf{Text identification}. $\mathcal{M}$ is said to identify $T$ just 
        in case there is a $p$ such that $\mathcal{M}(T)\converges = p$ and $W_p = \content(T)$, 
        i.e., just in case $\mathcal{M}$ conjectures the language $L = \content(T)$ on all 
        but finitely many prefixes of $T$.\label{def:text_indentification}
        
        \item \textbf{Language identification}. $\mathcal{M}$ is said to 
        identify the language $L \in \mathcal{E}$ just in case $\mathcal{M}$ 
        identifies every text for $L$.\label{def:language_indentification}
        
        \item \textbf{Identification of a classes of languages}. 
        Let $\mathcal{L} \subseteq \mathcal{E}$ be given. $\mathcal{M}$ identifies 
        $\mathcal{L}$ if and only if it identifies all $L \in \mathcal{L}$. Then,
        $\mathcal{L}$ is said \textbf{identifiable} in case there is a scientist 
        that identifies it; it is said to be \textbf{unidentifiable} otherwise. \label{def:identification_class_languages}
        
    \end{deflist}
    
\end{definition}

Several critical observations regarding this definition of identification need to be considered. 
Firstly, it is crucial to notice that, according to the definition, to identify a language, $L$, 
the scientist needs only to identify all the texts for the given language, without restrictions about what happens in other texts. 
Thus, in \gls{flt}, only the identifiability of classes of languages is worth discussing, 
since discussing the identifiability of singular languages is rendered superfluous, 
as every given single language $L \in \mathcal{E}$ is trivially identified by at least one scientist: the scientist 
that regardless of its input constantly outputs $p \in \mathbb{P}$, such that $p$ is the smallest grammar for $L$. 
We will name such a scientist a dumb visionary (\cref{def:dumb_visionary}).
Such broad definitions expand the ideas of both ``learner'' and ``learning'' to include these very dumb or stubborn behaviours, which becomes very useful to prove results but obviously raises criticism against this framework in what concerns its fidelity and capacity to model real learning behaviour. 
The field addresses these criticisms by extending the framework, redefining one or more of these components while being guided by already proven results. 

\begin{definition}[Dumb Visionary]\label{def:dumb_visionary} Given a language $L$, the dumb visionary for $L$, written $DV_L$, is defined as the scientist 
that, regardless of its input, constantly outputs the minimum program $p \in \mathbb{P}$ such that $W_p = L$. 
\end{definition}

One simple interesting result from the \gls{flt} field is that scientists that can be agnostic about the time they have converged, i.e., they are not imposed to announce their convergence, can identify more classes of languages than those imposed to announce their convergence. 
This and other intriguing results could be proved within this framework, yet in this text, we need to move on to discussing this theory inside the context of \gls{cc}. In this new context, the shop assistant can be replaced by a professional composer, yet no consequences of this change will actually be explored in this early work.

\section{Towards Formal Creativity Theory} \label{sec:creativity}

The definitions previously discussed were initially developed focusing on the learning component of language and science. 
We believe that these definitions can be refurbished to provide a more rigorous view on \gls{cc} concepts and possibly a new formalization to explore creativity in a purely theoretical way. 
Thus, the initial goal of this section is to examine some creativity-related terms through the lens of the previously provided definitions. 
We shortly introduce some concepts from the canonical literature of \gls{cc}, and what we consider could be their formal reinterpreted definition under the lens of \gls{flt}.  
Some of these terms are essentially renames of structures from \gls{flt}. We believe that the process of renaming has been instrumental in generating new concepts since it facilitates a clearer understanding.
We think of the set of these definitions as the beginning of a field that, by parallelism, we designate \gls{fct}. 

\subsection{Artefacts}

The seminal work by \citet{Rhodes1961-go}, proposing the four P's of creativity, introduces the idea of  `Product,' also traditionally referred to as artefact, as one of the main components of creative behaviour. For our purposes, we conceptualize an artefact as being represented by a sentence, or, equivalently, we redefine $\mathbb{A}$ as the set of all possible artefacts. This allows us to abstract and generalize our results across domains, but also pushes this theory away from practical research.

\begin{definition}[Artefact]
     A product, an artefact (or a concept) is any member $a \in \mathbb{A}$, the set of all possible artefacts. \label{def:artefact}
\end{definition}

\subsection{Experience and Fate Sequences}

While exploring the concept of `prefix' in the context of a creative scenario, it corresponds to the experience that an agent has access to in a given moment. Therefore, in this context, we decided to provide another term to refer to this structure: `experience sequence'.  This refers to the cumulative and ordered finite list of an individual's prior singular experiences with various artefacts. If we have an infinite list, i.e., a text, as we call it in \gls{flt}, we will call it a `fate sequence' in \gls{fct}. While a scientist can undergo an experience sequence, they can never totally undergo a fate sequence.

\begin{definition}[Experience and Fate Sequences]
    
    An experience (sequence) is a prefix, $\sigma \in SEQ$, that can be undergone by scientist. A fate (sequence) is the same as a text, $T$. \label{def:experience}

\end{definition}

\subsection{Situation}

We decided then to define `situation' as a pair consisting of a scientist and an experience. Despite not being a common idea in \gls{cc}, we believe that formalizing this component as a pair is crucial for later analysis, especially when considering properties of an artefact that depend on the experience, the scientist, or the interaction between the two.

\begin{definition}[Situation]
     A situation $s$ is any ordered pair $s = (\mathcal{M}, \sigma)$ with $\mathcal{M}$ a scientist and $\sigma$ an experience. We use $\mathbb{S}$ to represent the class of all situations. \label{def:situation}
\end{definition}

\subsection{Conceptual Spaces}

Conceptual spaces have been commonly discussed in \gls{cc} \citep{Boden2004-ha,Gardenfors2004-pv}, even though sometimes with different meanings.
We propose to formalize a conceptual space as a \gls{re} set, i.e., any set of artefacts that can be intentionally defined (\cref{def:conceptual_space}).
This way, any program over artefacts defines a conceptual space and all of these can be intentionally defined using a hypothesis.
This definition aligns with \citepos{Ritchie2006-ye} idea of conceptual space\footnote{  \citet{Ritchie2006-ye} suggests that the term `artefact-set' corresponds to the set that actually contains artefacts, which we simply call the conceptual space, and the term `space-definition' to the intentional definition of this set, which we call the hypotheses or the rules of the conceptual space, as mentioned later in this text.}. Furthermore, we wish to distinguish three special kinds of conceptual space: the inspiring set, the hypothetical conceptual space, and the platonic conceptual space.

\begin{definition}[Conceptual Space]
    A conceptual space is any \gls{re} set, $A \in \mathcal{E}$. \label{def:conceptual_space}
\end{definition}

\subsubsection{Inspiring Sets}

According to \citet[p.~76]{Ritchie2007-ii}, in \gls{cc} the ``construction of the program is influenced (either explicitly or implicitly) by some subset of the available basic items (...) which we will call the inspiring set.'' 
Following the definitions from \gls{flt}, it seems logical to define the inspiring set based on what is available to the scientist, i.e., experiences. Then, we have that the inspiring set is precisely the unordered content of the experience sequence (\cref{def:experience}). The set and the sequence are then two different ways to represent inspiration, the latter allowing repetitions, while the former does not. 

\begin{definition}[Inspiring Set]
     
     Let $\sigma$ be an experience, then we also call $\content(\sigma)$, the inspiring set, written $I_\sigma$ or just $I$ when the prefix is not ambiguous. \label{def:inspiring_set}
\end{definition}

\subsubsection{Hypothetical Conceptual Spaces}

This concept is not commonly found in \gls{cc} literature. Instead, it resulted from the exploration of creativity through the lens of \gls{flt}.
A hypothetical conceptual space, $H$, is any conceptual space conjectured by a scientist (\cref{def:hypothetical_conceptual_space}). 
We see then that a situation $s \in \mathbb{S}$ totally defines a hypothetical conceptual space\footnote{Non-total scientists that are not defined in all experiences are possible, allowing to model indecision but adding extra complexity.}. 
Additionally, given a fate, we have an infinite sequence of hypothetical conceptual spaces.

\begin{definition}[Hypothetical Conceptual Spaces]
     Let $s=(\mathcal{M},\sigma)$ be a situation, then we say that $H_s = W_{\mathcal{M}(\sigma)}$ is the hypothetical conceptual space associated with that situation, i.e., it is the conceptual space conjectured by scientist $\mathcal{M}$ when undergoing $\sigma$. \label{def:hypothetical_conceptual_space}
\end{definition}

In \gls{flt}, there is a branch dedicated to studying the implications of assuming certain relationships between the hypothetical conceptual spaces and the corresponding inspiring sets. 
For example, a consistent scientist is any scientist such that for any situation $s =(\mathcal{M}, \sigma)$ we have that $I_\sigma \subseteq H_s$\footnote{These and other kinds of constraints imposed on scientists are considered a type extension to the \gls{silitf} known as strategies.}.

\subsubsection{Platonic Conceptual Space}

The platonic conceptual space, $P$, is the correct reality among the possible ones, the conceptual space that originates a given fate. Conceptually, changing the platonic conceptual space means considering an alternative reality. 
The fate, the infinite sequence of artefacts the scientist will experience, is always congruent with a given platonic conceptual space and contains all of its elements\footnote{\Gls{flt} includes other structures that consider other relationships between $P$ and $T$ - a type of extensions named environments - accounting for missing data, wrong data, or additional information.}. The fate, then, represents the sequence that encompasses all the artefacts the scientist will ever have access to while trying to guess $P$.

\begin{definition}[Platonic Conceptual Space]
    Given a fate $T$, the platonic conceptual space, $P$, with which $T$ is congruent, is such that $P = \content(T)$.  \label{def:platonic_conceptual_set}
\end{definition}
   
Paraphrasing the previously defined criterion for identification (\cref{def:text_indentification}), a scientist is said to identify a fate, $T$, if and only if in all but finitely many experiences of $T$, the hypothetical conceptual space is the same as the platonic conceptual space\footnote{The last kind of extension is to consider alternative success criteria to simulate approximate learning or indecision.}. Also, a collection of platonic conceptual spaces is identifiable if, for each one $P$ of these, there is a scientist capable of identifying all fates congruent with $P$.  
In the anecdote, the platonic conceptual space corresponds to the client's idea of `grey music' and the hypothetical conceptual spaces to the several hypothetical genres the assistant goes through after each provided song.

\subsection{Rules}

In the \gls{csf}, \citet{Wiggins2019-kv} presents three sets of rules  - $\mathcal{R}$, $\mathcal{T}$, $\mathcal{E}$. 
These rules play different roles in the creative process - the first one is used to define the boundaries of a special conceptual space, the second is used to generate new artefacts and the last one is used to choose the artefacts with value. While both the first and third are used to distinguish artefacts that belong to a specific conceptual space or those that are valuable, respectively, the second one provides an ordered sequence of artefacts. In \gls{silitf}, the notion of hypothesis corresponds to the implicit representation of a language. Seems appropriate, then, to correspond such sets of rules to hypotheses  (\cref{def:rules}), despite some being used to decide and others to enumerate.
In the future, we intend to explore  \gls{csf}-aligned scientists, that output 3 sets of rules, $p_R$, $p_T$ or $p_E$, instead of just one. 

\begin{definition}[Rules]
     A set of rules is a program, $p \in \mathbb{P}$. \label{def:rules}
\end{definition}

\subsection{Properties}

In traditional \gls{flt}, various extended models for data (environments), classes of learners (strategies), and success criteria (criteria) are explored. 
Yet, to our knowledge, there is close to no exploration of how these can take advantage of the properties of the artefacts, perhaps due to abstraction complexity. 
Conversely, in \gls{cc} literature, it is a common but not exclusive viewpoint that a creative act is characterized by a process yielding creative artefacts. 
Therefore, many of the efforts in \gls{cc} wander around artefacts and their properties.
Several have then been the properties proposed as desiderata to discern creative artefacts, such as novelty, typicality, value, quality or even surprise.

We propose to include `property extensions' to \gls{fct} as a new extension type.
This new kind of extension enables the creation of new paradigms, by constraining the class of scientists based on these properties, or just by employing different criteria that define success based on such properties. 
Due to space limitations, we do not fully detail how properties can alter the basic paradigm but will briefly introduce two formal properties and some conclusions.

\subsubsection{Schema}  
To implement this extension, we adapt the idea of `rating schemas', proposed by \citet{Ritchie2007-ii}, that correspond to operators that can be applied to artefacts. 
While \cite{Ritchie2007-ii} implements schema ratings as fuzzy sets, for simplicity, in this initial approach to \gls{cc} using \gls{flt}, we suggest to have schemas as predicates (\cref{def:schema}).
Similarly to \citeposs{Wiggins2006-qy} dichotomic evaluation rules, $\mathcal{E}$\footnote{Though Wiggins focuses on predicates, he mentions in the paper the potential for extending his proposal to fuzzy sets.}, this simple implementation allows for only two states of value: valuable artefacts and non-valuable artefacts. 
Despite facilitating theory, this simplicity-driven design choice can raise criticism, since our findings may not apply to alternative rating schema models that might be more accurate. 
However, it aligns with \gls{flt} methodology: proving results for a simplified model first, then extending to more complex versions, with insights from the simpler approach facilitating subsequent proofs.

\begin{definition}[Schema]
     A (dichotomic rating) schema $V$ is just any predicate over artefacts, $V: \mathbb{A} \rightarrow \{0,1\}$. An artefact $a$ is then said to be $V$-valuable if $V(a) = 1$. \label{def:schema}
\end{definition}


According to this definition, any predicate over artefacts can be interpreted as a schema, regardless of the domain, allowing for uncomputable schemas that can never be completely correctly grasped in a conceptual space. 
Yet, a set of rules for a conceptual space can always be used to build a schema. 
In this study, we decided to focus on schemas parameterized by an experience and a scientist. 
We provide next the working definitions of one well-known and widely used property of artefacts from \gls{cc} - novelty, $V_N$ - as well as introduce a new one - transformativeness, $V_{Tr}$ - relating to transformational creativity \citep{Boden2004-ha}.

\subsubsection{Novelty}

While properties such as complexity or typicality have been debated as necessary components for creativity, novelty has increasingly been agreed as one main one, to the point that creativity has been seen as the ``production of novel,  useful products'' \citep[p.~110]{Mumford2003-wh}. \citet[p.~25]{Wiggins2006-qy} defines novelty as ``the property of an artefact (abstract or concrete) output by a creative system which arises from prior non-existence of like or identical artefacts in the context in which the artefact is produced.'' We define that an artefact $a$ is novel, given an inspiring set $I_\sigma$, if $a \notin I_\sigma$. Despite this being a reductionist notion of novelty, it allows to derive interesting results when paired with other properties. 

\begin{definition}[Novelty]
     Novelty $V_N$ is a schema parameterized by a situation $s=(\mathcal{M}, \sigma)$. An artefact $a$ is said to be \textit{novel} in situation $s$, writing $V_N(a,s) = 1$, iff $a \notin I_\sigma$. \label{def:novelty}
\end{definition}

We can see that a scientist can compute this value metric since it only depends on the experience. This means that the scientist can access the true value of this schema. This is not true for all schemas, as there are schemas that can only be approximated by the scientists, which will be the case for the next schema. Since our model of these schemas is parameterized, we can see that novelty is not intrinsic to an artefact, it depends on the experience provided to a scientist. It makes no sense then to evaluate this predicate out of the context of a specific situation.

\subsubsection{Transformativeness}

Following \citepos{Boden2004-ha} proposal of different kinds of creativity - exploratory, combinational and transformational creativity - more common than not, research in \gls{cc} is influenced by the ``widely discussed proposal by Boden that high levels of creativity result from the transformation of a conceptual space'' \citep[p.~242]{Ritchie2006-ye}.  \citet[p.~31]{Wiggins2019-kv} paraphrases Boden's transformational creativity as ``the kind of creative behaviour concerned (\ldots) with transforming the rule set'' that defines a conceptual space.

The interplay between transformational creativity and learning is inherently complex and deeply interconnected. 
Through analyzing hypothetical scenarios of creative behaviour, we identified two main cases for the relationship between learning and creativity: 
one where learning precedes exploration, akin to the preparation phase described by \citet{Wallas1926-og};  
and another where learning occurs during exploration, aligning with what \cite{Wiggins2019-kv} describes as transformational creativity. 
When analyzing real cases, distinguishing when exploration begins and preparation ends become subjective, so one can only differentiate transformational creativity from well-prepared exploratory creativity by fixing a starting point for the exploration process.  

Generalizing \citeposs{Wiggins2019-kv} idea of transformational creativity as changing an intentional representation, and relating it to how scientists learn in \gls{flt}, we say that transformational creativity happens whenever a scientist changes their hypothesis in a growing experience. 
Formally, we observe transformational creativity if a scientist outputs $h_1$ from experience $\sigma_1$, and then generates a different hypothesis $h_2$ when undergoing $\sigma_1\diamond a$. 
Not all artefacts will trigger such shift in a specific situation. Those which do are said transformative, a property we named `transformativeness'.

\begin{definition}[Transformativeness]
     Transformativeness $V_{Tr}$ is a schema parameterized by a situation $s=(\mathcal{M}, \sigma)$. An artefact $a$ is said to be transformative in situation $s$, i.e., $V_{Tr}(a,s) = 1$, iff $\mathcal{M}(\sigma) \neq \mathcal{M}(\sigma \diamond a)$. \label{def:transformativeness}
\end{definition}

This is, to the best of our knowledge, the first formal proposal to define transformational creativity that allows to directly compare with other properties such as novelty. 

\section{Preliminary Results}

Novelty has been widely discussed in Computational Creativity as a mandatory requirement for creative behaviour. The common way to study this aspect then became to study what makes a product novel and find search strategies that could increase the likelihood of novelty. 
Yet, one other aspect of novelty we can study, given these new definitions, is its relationship with other properties such as transformativeness. We aim to study how experiencing novel artefacts affects learning or, in our case, how it affects creativity - namely how necessary it is for transformational creativity \citep{Boden2004-ha}. It is unspokenly assumed that novelty is needed for transformational creativity to happen. This is yet another reason for novelty being valued in creativity - because novel artefacts are assumed to impact conceptual models. We prove here that if we consider the general class of scientists, it is wrong to assume that there is a relationship between novelty and the transformation of conceptual spaces! We prove that, in this wide class of agents, there is no such relationship and we illustrate the results using examples. We then show that for a sub-class of scientists, it is correct to assume such a relationship.

\subsection{Novelty is not sufficient for Transformativeness}

Could it be that all pieces of novel information will always make one change their mind? In this framework, we can define scientists that defy such assumption. Actually, we have infinitely many scientists that break such statement, some very trivial, others more complex.

\begin{theorem}\label{thm:nov_not_sufficient_for_transformativeness}
    Artefact $a$ being novel in situation $s$ does not imply $a$ is transformative in that same situation.
\end{theorem}

\begin{proofs}[thm:nov_not_sufficient_for_transformativeness]
    We will provide a situation $s$ and an artefact $a$ such that $a$ is novel but not transformative in situation $s$.
    Suppose $M_h$ is a constant scientist, i.e., there is constant $h \in \mathbb{N}$, such that for every $\sigma \in SEQ$,  $M(\sigma) = h$. An example of such kind of scientists is what we defined as dumb visionaries $DV_A$ (\cref{def:dumb_visionary}). Now, suppose a situation $s = (M_h, \sigma)$ with a fixed $\sigma \in SEQ$. Suppose as well any artefact $a$ such that $a \notin I_\sigma$. Such $a$ must exist since $I_\sigma$ is always finite due to the finitude of $\sigma$, i.e., independently of the chosen experience there is always at least one novel artefact. By definition, we see that the chosen $a$ is novel in $s$, yet $M(\sigma)=h=M(\sigma \diamond a)$, hence $a$ is not transformative.
\end{proofs}

As this proof shows, we showcase scientists for whom novelty alone is insufficient to alter their ideas. Even though these seem very simple scientists, we can build more complex scientists that would prove such a theorem, such as any scientist who identifies an infinite language. 
If novelty was sufficient to make any scientist change their opinion, then any fate for an infinite platonic conceptual space would have infinite points of transformation and therefore no scientist could ever identify it. 
Yet, we think the simple scientists provided in the proof illustrate well the argument in point, since despite their triviality these simple scientists still identify one platonic conceptual space. 
We can compare these to people with unjustified beliefs who are immune to any kind of evidence, yet can be, by chance, right in some cases/realities, and that despite novelty they will never change their beliefs.

\subsection{Novelty is not necessary for Transformativeness}

Maybe even more surprising is that, within these definitions, novelty is also not required for transformativeness. Transformativeness can occur in situations and artefacts that fail to meet the novelty criterion. 

\begin{theorem}\label{thm:nov_not_necessary_for_transformativeness}
    Artefact $a$ being transformative in situation $s$ does not imply $a$ is novel in that same situation.
\end{theorem}

\begin{proofs}[thm:nov_not_necessary_for_transformativeness]
    In this case, we need to provide a situation $s$ and an artefact $a$ such that $a$ is transformative but not novel in situation $s$.
    Suppose $M'$ is an ever-changing scientist, i.e., $M'(\sigma) \neq  M'(\sigma \diamond b)$ for every $\sigma$ and $b$. An example of such a scientist is a scientist whose output depends only on the size of the experience, such as $M'(\sigma) = |\sigma|$
    Suppose a situation $s = (M', \sigma)$ for fixed $\sigma \in SEQ$ but $I_\sigma \neq \emptyset$. Let us choose an artefact such that $a \in I_\sigma$.
    We see that the chosen $a$ is not novel, because we explicitly chose something that was already experienced by the scientist, yet our scientist changes always their hypothesis when his experience grows so $M(\sigma) \neq M(\sigma \diamond a)$ hence $a$ is transformative.
\end{proofs}

Once again, these are very trivial scientists. These, contrary to the constant ones, are actually not able to identify anything. A more complex and more realistic example to prove such a result would be a scientist that annotates how confident it is in its hypotheses by appending a comment to the hypotheses. If the artefact matched its hypotheses then it would increase the confidence by one point, otherwise it would decrease the confidence or even make it change completely if confidence reached zero. Such scientist is by definition ever-changing and therefore also proves the theorem above, for syntactic transformativeness. This scientist can converge semantically and can be used to identify using an expansion to the identification criterion previously presented - behaviourally correct identification.

\subsection{For Set-Driven Scientists, Novelty is necessary for Transformativeness}

A scientist is said to be set-driven if for $\sigma, \tau \in SEQ$, if  $\text{content}(\sigma) = \text{content}(\tau)$  then $M(\sigma) = M(\tau)$. A ``set-driven scientist is insensitive to the order in which data are present'' \cite[p.~99]{Jain1999-gh}. We will prove that for these scientists, novelty is a requirement for transformativeness. We have two conditions:
\begin{enumerate}
    \item \textbf{Condition 1}: for scientist $M$ and $\sigma,\tau \in SEQ$, $\content(\sigma) = \content(\tau) \implies M(\sigma) = M(\tau)$, i.e., using inspiring set notation, $I_\sigma = I_\tau \implies M(\sigma) = M(\tau)$
    \item \textbf{Condition 2}: for artefact $a$ and situation $s=(M,\sigma)$, $V_{Tr}(a,s) \implies V_N(a,s)$ 
\end{enumerate}

\begin{theorem}\label{thm:nov_necessary_for_transformativeness_set_driven}
    \textbf{Condition 1} is sufficient for \textbf{Condition 2}, i.e., if a scientist is set-driven then novelty is necessary for transformativeness in that same situation.
\end{theorem}

\begin{proofs}[thm:nov_necessary_for_transformativeness_set_driven]
    Let us start by assuming \textbf{Condition 1}, i.e., let  $M$ be a set-driven scientist. So given any artifact $a$, and any experience $\sigma$, we have $I_\sigma =  I_{\sigma \diamond a} \implies M(\sigma) = M(\sigma \diamond a)$. 
    This means the contrapositive $M(\sigma) \neq M(\sigma \diamond a) \implies I_\sigma \neq I_{\sigma \diamond a}$. 
    Thus for any situation $s=(M,\sigma)$, we have then that $V_{Tr}(a,s) \implies I_\sigma \neq I_{\sigma \diamond a}$. 
    In turn, $I_\sigma \neq I_{\sigma \diamond a} \implies (I_\sigma - I_{\sigma \diamond a} \neq \emptyset \lor I_{\sigma \diamond a} - I_\sigma \neq \emptyset)$. 
    Additionally to the previous one, we prove now two sub-results. 
    A) From the definitions of inspiring set and $\diamond$, we know that for any $\sigma, \tau$  we have that $I_\sigma \subseteq I_{\sigma \diamond \tau}$, then, we always have also that $I_\sigma - I_{\sigma \diamond \tau} = \emptyset$. 
    B) Similarly, for any $\sigma, \tau \in SEQ$, it is always true that $I_{\sigma \diamond \tau} - I_{\sigma} \subseteq I_\tau$,  thus $I_{\sigma \diamond a} - I_\sigma \subseteq \{a\}$.
    Given the last three results, we may conclude that $I_{\sigma \diamond a} - I_\sigma = \{a\}$, and that therefore $a \notin I_\sigma$, that corresponds to our definition of novelty, $V_N(a,s)$, so we reached \textbf{Condition 2}.
\end{proofs}

The consequences of this theorem are significant. At first glance, we see that it is appropriate to assume that novelty is actually necessary for transformational creativity when considering set-driven agents such as those based on the inspiring sets proposed by \cite{Ritchie2007-ii}. 
Yet, on a second look we may also conclude that assuming such set-driven properties seems to constitute a big change in the paradigm, implicating several other assumptions and changing drastically the way creativity is modelled.
We believe it is of the most relevance to understand the implications and consequences of such assumptions. 
This can only be achieved by formal studies and bringing clarity to how these assumptions relate as well as their limitations. 
Implementation of set-driven systems can benefit from a better knowledge of these assumptions, since human behaviour seems sensitive to repetition and order in experiences. 

Note that the presented result does not mean that whenever we consider novelty as necessary for transformativeness we are actually considering set-driven scientists. We can use the formal tools to prove that this is not the case. As a short version of this proof, we can suggest the case of the scientist who always outputs the last novel artefact, which meets \textbf{Condition 2} while not being set-driven (not \textbf{Condition 1}).

\section{Future Work}

The richness of this research path  made the space restrictions of this format quite challenging. Thus, there are several promising directions that stayed out of this paper.  
Firstly, alternative definitions, such as semantic transformativeness, and the formalization of other \gls{cc} concepts, like typicality, both represent opportunities to enhance our understanding of creativity's subtleties. 
Additionally, we also aim to uncover new insights through the formal modeling of other assumptions, such as \citeposs{Wiggins2006-qy} premise that transformative artifacts must lie outside the current hypothetical conceptual space — termed aberrations. 
Another avenue involves considering schemas parameterized by various components, such as fates (\cref{def:experience}), or moving beyond predicates to embrace alternative schema representations like fuzzy sets or semi-orders. 
Lastly, a critical exploration into the computability of these properties could yield valuable knowledge about the practicality and limitations of applying these theoretical constructs in real systems.

Despite mentioning `property extensions' within this text, we have only defined some of the properties that could underpin a new paradigm, never explaining how we could build such paradigm out of these definitions.
A longer format would be necessary to fully present these paradigms, what they can model and how they relate to other classic \gls{flt} extensions such as accountable scientists, incomplete fates and the behaviourly correct criterion.

Lastly, a primary objective for our future research is to explore scientists capable of producing multiple sets of rules, reviewing the role discrimination and generation have within theory of computation. This approach aims to more accurately model the \gls{csf}, providing a nuanced understanding of creativity through the lens of FLT.

\section{Conclusion}

According to \citet[p.~36]{Wiggins2019-kv}, a ``theoretical framework, such as that presented here, may be useful in teasing out philosophical issues, but it may also be useful in giving generalised descriptions of behaviours which might be observed in creative agents.'' The present paper provides an accessible way to understand the advantage of using theoretical frameworks -- namely \gls{flt} -- to study \gls{cc}. 

In this paper, we prove that novel artefacts are neither necessary nor sufficient for transformational creativity when we consider the wide range of possible behaviours over a sequence of artefacts. However, if we consider agents that disregard the order as well as the repetitions, then novelty becomes a requirement. We concluded then that agents based only on inspiring sets, as suggested by \cite{Ritchie2007-ii}, might be disregarding some of the characteristics that make human creativity interesting, such as the fact that the same artefact can actually change one's mind more than once.

While practical research is crucial, its overvaluation often sidelines theoretical studies. The current trend of valuing research topics mainly for their immediate practical applications can, if overemphasized, risk limiting broader societal progress to short-term gains. Theory can clarify the consequences of certain assumptions and guide developments in the long run. It is necessary to revitalize fundamental research by providing novel, pure, and non-applied results.

These approaches are one of the characteristics that distinguishes what we consider to be \gls{cc} and Creative AI so we believe it is time to finally make it official and have a sub-branch dedicated to these matters -- the sub-branch of \gls{cct}. 
We believe this suggestion aligns well with \citepos{Colton2015-vw} original proposal of \gls{cct} that aimed at establishing a theory in \gls{cc} akin to \gls{clt}. However, it is worth noting that over time this term has more often than not been associated with specific models -- FACE and IDEA \citep{Pease2011-py,Colton2011-rt} -- published by the same authors under this banner. Our intention is to reclaim and broaden the term to designate a comprehensive research area, which closely aligns with what we think was \citepos{Colton2015-vw} original aspiration for \gls{cct} and possibly fulfilling its original aim. To that end, we include under \gls{cct}, the new proposed set of definitions -- the beggining of \gls{fct}. 

\section{Author Contributions}
The primary author took the lead in writing and developing the theoretical framework of this paper as part of their PhD work, which included the formulation of theorems and their respective proofs. The co-authors, who are the primary author's supervisors, made substantial contributions through their critical review, feedback, and engaging discussions during supervisory meetings. Their expertise and insights were crucial in refining the definitions, enhancing the presentation of the matters, and ensuring the scholarly quality of the final manuscript.

\section{Acknowledgments}
We extend our heartfelt thanks to our colleagues at the CCLab and the CMS, with special appreciation for André Santos and Márcio Lima, for their invaluable and constant support. This work is funded by national funds through the FCT - Foundation for Science and Technology, I.P., within the scope of the project CISUC - UID/CEC/00326/2020 and by European Social Fund, through the Regional Operational Program Centro 2020. The first author is also individually funded by Foundation for Science and Technology (FCT), Portugal, under the PhD fellowship grants (2021.05529.BD).

\bibliographystyle{iccc}
\bibliography{iccc}

\end{document}